\documentclass{article}
\newcommand\tab[1][1cm]{\hspace*{#1}}

\usepackage{arxiv}

\usepackage{subfig}

\usepackage[utf8]{inputenc} 
\usepackage[T1]{fontenc}    
\usepackage{hyperref}       
\usepackage{url}            
\usepackage{booktabs}       
\usepackage{amsfonts}       
\usepackage{nicefrac}       
\usepackage{microtype}      
\usepackage{lipsum}
\usepackage{mathtools}
\usepackage{mathrsfs}
\usepackage{amsmath}
\usepackage{array}
\usepackage{algorithm}
\usepackage[noend]{algpseudocode}
\usepackage{graphicx}
\usepackage{bm}
\usepackage[most]{tcolorbox}
\usepackage[justification=centering]{caption}
\usepackage{caption}
\usepackage{xcolor}
\usepackage[margins, adjustmargins]{trackchanges}
\addeditor{AB}
\usepackage[normalem]{ulem}
\usepackage{caption}
\usepackage{multirow}
\usepackage{booktabs}
\usepackage{makecell}
\usepackage{tikz}\usetikzlibrary{positioning,angles,quotes,calc,arrows,arrows.meta,positioning, decorations.pathreplacing}
\algnewcommand\algorithmicinput{\textbf{Input:}}
\algnewcommand\algorithmicoutput{\textbf{Output:}}
\algnewcommand\Input{\item[\algorithmicinput]}%
\algnewcommand\Output{\item[\algorithmicoutput]}%
\usepackage{marginnote}
\usepackage{soul}

\graphicspath{ {./images/} }

\newcommand{\bc}{\begin{center}}
\newcommand{\ec}{\end{center}}
\newcommand{\bea}{\begin{eqnarray}}
\newcommand{\eea}{\end{eqnarray}}

\makeatletter
\newcommand*{\rom}[1]{\expandafter\@slowromancap\romannumeral #1@}

\newcolumntype{M}[1]{>{\centering\arraybackslash}m{#1}}

\newcolumntype{C}{>{\centering\arraybackslash}m{8em}}

\makeatletter
\def\BState{\State\hskip-\ALG@thistlm}
\makeatother

\title{Curriculum Learning with Hindsight Experience Replay for Sequential Object Manipulation Tasks}
\author{
	Binyamin Manela \tab Armin Biess \\
	manelab@post.bgu.ac.il \tab abiess@bgu.ac.il \\
	Department of Industrial Engineering and Management\\
	Ben-Gurion University of the Negev\\
	Beer-Sheva, Israel \\
}

\begin{document}
	
\maketitle
	
\begin{abstract}
Learning complex tasks from scratch is challenging and often impossible for humans as well as for artificial agents. A curriculum can be used instead, which decomposes a complex task (target task) into a sequence of source tasks (the curriculum). Each source task is a simplified version of the next source task with increasing complexity. Learning then occurs gradually by training on each source task while using knowledge from the curriculum's prior source tasks. In this study, we present a new algorithm that combines curriculum learning with \textit{Hindsight Experience Replay} (HER), to learn sequential object manipulation tasks for multiple goals and sparse feedback. The algorithm exploits the recurrent structure inherent in many object manipulation tasks and implements the entire learning process in the original simulation without adjusting it to each source task. We have tested our algorithm on three challenging throwing tasks and show vast improvements compared to vanilla-HER.
\end{abstract}


\section{Introduction}
Deep reinforcement learning, i.e., the combination of reinforcement learning \cite{sutton2018reinforcement} with deep learning \cite{goodfellow2016deep}, has led to many breakthroughs in recent years for generating goal-directed behavior in artificial agents, ranging from playing Atari games with superhuman capabilities \cite{mnih2013playing} to making animated figures walk \cite{todorov2012mujoco, lillicrap2015continuous, schulman2017proximal}, solving complex manipulation tasks \cite{popov2017dataefficient, openai2018learningdexterous}, and defeating the best GO player in the world \cite{silver2018general}. All reinforcement learning problems are based on the reward hypothesis, stating that any goal-directed task can be formulated in terms of a reward function. However, the engineering of such a reward function is often challenging and limits the application of reinforcement learning to real-world tasks, for example, in robotics \cite{kober2013reinforcement}. Hindsight Experience Replay (HER) \cite{andrychowicz2017hindsight} circumvents reward engineering by using sparse reward signals to indicate whether a task has been completed or not. HER uses failures to learn how to achieve alternative (virtual) goals that have been achieved in the episode and uses the latter to generalize to actual goals. Furthermore, HER can target multiple goals and provides a form of curriculum learning. Goals that can be easily achieved, for example, by sampling from a trajectory using a random policy, are learned first and then used in replay for the learning of more difficult goals. Although HER showed excellent performance for simple tasks (e.g., \textit{pushing}, \textit{sliding} and a \textit{simplified pick-and-place} task), it fails in complex object manipulation tasks, which are composed out of a sequence of sub-tasks.
For example, consider the \textit{pick-and-place} task of a box. HER will make little progress towards the goal since the agent needs first to solve the \textit{pick} task before it can proceed to the \textit{place} task. Therefore, in the original HER paper, the authors used a \textit{simplified pick-and-place} task, where the box initialized in the gripper half of the time. In this paper, we try to overcome these limitations by using HER within a curriculum learning approach. Our main contributions are as follows: First, we provide a curriculum learning algorithm based on HER for sequential object manipulation tasks. The curriculum is derived from the inherent recurrent structure of the task.
Second, we design a neural network architecture that is particularly adapted to sequential object manipulation tasks and show that the algorithm reduces exponential run-time of HER to polynomial time using a curriculum. Third, we test our algorithm on three different \textit{throwing} tasks with varying levels of complexity.

\section{Background}
In this section, we provide background information for reinforcement learning, hindsight experience replay, and curriculum learning.

\subsection{Reinforcement Learning}
Reinforcement learning aims to solve a sequential decision-making process by trial-and-error while interacting with an environment \cite{sutton2018reinforcement}. Generally, we assume that the environment is fully observable and defined by a \textit{Markov Decision Process} (MDP) with a set of states $s \in \mathcal{S}$, set of actions $a \in \mathcal{A}(s)$, initial state distribution $P(s_0)$, reward function $r: \mathcal{S} \times \mathcal{A} \rightarrow \mathbb{R}$ and a discount factor $\gamma \in [0,1]$. The agent acts in the environment according to a policy $\pi$ that maps states $s$ to actions $a$. At the beginning of each episode, an initial state $s_0$ is sampled from the distribution $p(s_0)$. At every timestep, the agent chooses an action $a_t$ according to policy $\pi(s_t)$, performs this action and receives an immediate reward and the next state from the environment. The episode is terminated when the agent reaches a terminal state or exceeds the maximum number of timesteps. The agent's goal is to find the policy that maximizes expected return, i.e., cumulative future discounted reward $R_t=\sum_{i=t}^{\infty}\gamma^{i-t}r_i$.
	
\subsection{Hindsight Experience Replay (HER)}
The traditional RL algorithm can be extended to multi-goal tasks using \textit{Universal Value Function Approximators} (UVFA) \cite{schaul2015universal}. The key idea behind UVFA is to augment action-value functions and policies by goal states, and thus, every transition contains also the desired goal. This enables generalization not only over states but also over goals when using neural networks as function approximators. 
In multi-goal tasks with \textit{sparse} rewards, it is challenging to achieve progress and learn a task. HER addresses this problem by taking failure as a success to an alternative (or \textit{virtual}) goal. HER applies UVFA and includes new transitions with virtual goals. Thus, the agent can learn from failures through generalization to actual goals. It has been demonstrated that HER significantly improves performances in various challenging simulated robotic environments. 
	
\subsection{Transfer learning}
Transfer learning can be used to facilitate learning when a task is challenging to learn \cite{lazaric2012transfer, taylor2009transfer}. After training on a \textit{source task}, learned knowledge could be used to initialize the agent in a second, related \textit{target task}. The objective of transfer learning is to transfer knowledge from the source task into the target task to improve performance, for example, by reducing the number of samples needed to learn a nearly-optimal solution. Transfer knowledge can be encoded in terms of value functions, policies, options, models, and even samples. The knowledge is usually transferred to the target task in a \textit{one-step} process from all source tasks. Source and target tasks may have different action and state spaces for which inter-task mappings need to be defined or learned. Transfer performance is measured by comparing the learner's performance with and without transfer using different metrics, such as time-to-threshold, jumpstart, asymptotic performance, total reward, and transfer ratio (ratio of total rewards with and without transfer).

\subsection{Curriculum learning}
Curriculum learning, also known as \emph{sequential transfer learning}, is an extension of transfer learning, where the agent is trained on a chosen \textit{sequence} of source tasks (i.e., a curriculum) $M_1, M_2, \dots M_t$, that is composed out of simplified versions of the target task. The goal in curriculum learning is to design a sequence of source tasks so that the agent's final performance or learning speed is improved. Transfer learning is used to transfer knowledge between each pair of tasks in the curriculum \cite{narvekar2017curriculum, bengio2009curriculum}. Strong transfer occurs if the training time spent to learn the target \textit{and} source tasks is improved when compared to learning the task from scratch. \\
In this paper, the source tasks and the curriculum are resulting from the inherent recurrent structure of manipulation tasks. All source tasks will have the same action spaces but will differ in state spaces, which will increase along the sequence of source tasks.

\section{Sequential-HER} \label{s:Sequential-HER}
In this section, we present our novel algorithm - \textit{Sequential-HER} ({SHER}) - which consists of two steps that are applied sequentially to each source task. First, we adapt and apply HER to the current source-task (section \ref{ss:Source-task adapted HER}). Second,  we transfer the policy learned in the current source task to the next source task (section \ref{ss:Knowledge transfer between sub-tasks}).

\subsection{Source-task adapted HER} \label{ss:Source-task adapted HER}
\begin{figure}
	\centering
	\subfloat[Source task $\psi_1$]
	{{\includegraphics[width=0.4\textwidth]{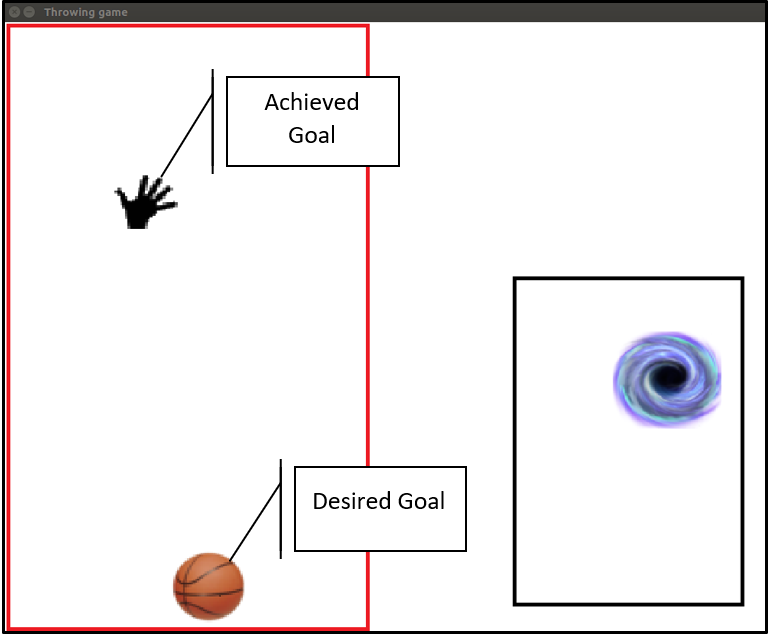} 
			\label{fig:subtask1}}}
	\hskip 3ex
	\subfloat[Source task $\psi_2$]
	{{\includegraphics[width=0.4\textwidth]{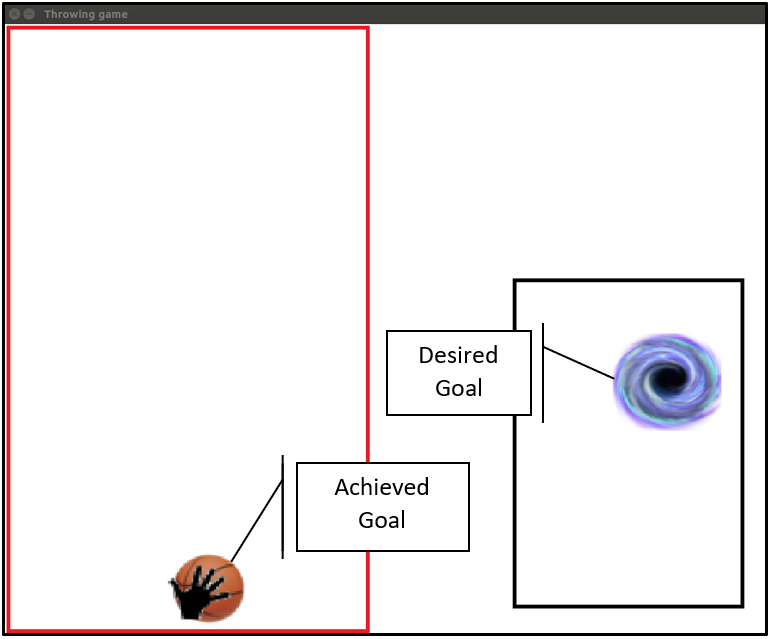} 
			\label{fig:subtask2}}}%
	\caption{An illustration of the recurrent structure of the \textit{Hand} task and its sequencing into two source tasks $\psi_1$ and $\psi_{2}$ (a and b). Our approach assumes that the desired goal for the object of source task $\psi_1$ is the achieved goal of the object in source task $\psi_{2}$.
	The red and black bounding boxes indicate the workspace of the	hand and the region of possible target locations, respectively. The objects' dimensions were enlarged for demonstration purposes.}
	\label{fig:subtasks}%
\end{figure}

Each source task consists of an observation dictionary of the same structure used in the original HER algorithm:
\begin{itemize}
    \setlength\itemsep{0.1em}
	\item \textbf{observation} - the current state of the environment
	\item \textbf{achieved goal} - the goal achieved in the current state
	\item \textbf{desired goal} - the goal of the current episode
\end{itemize}
The achieved goal is the position of the object in the current source task, and the desired goal is the desired position for the object. Our approach assumes that there is a physical interaction between the objects of source task $\psi_i$ and $\psi_{i+1}$. In other words, the desired goal for the object of source task $\psi_i$ is the object's location in source task $\psi_{i+1}$. Many common sequential manipulation tasks satisfy this condition, such as tool handling (hammering, screwing, drilling, etc.) and object manipulation (pushing, throwing, etc.). For some tasks, however, this condition is violated. For example, in a game, where an agent needs to press a button to unlock a ball and then pick and throw it towards a target, the object of source task $\psi_1$ is given by the button, whereas the object of source task $\psi_2$ is given by the ball. As there is no physical interaction between the two objects, our approach cannot be directly applied. Still, the former task, as well as other multi-object manipulation tasks, can be solved by breaking it down into a set of sequential tasks, where each sub-task (reach to the button, throwing of the ball) satisfies the above condition, and then learn these sub-tasks independently. Nonetheless, there are tasks for which the algorithm cannot be applied,  such as juggling, in which the objects (balls) cannot be distributed over different agents. \\
Due to the recurrent structure of our task, the state space, and the achieved- and desired goals can be predefined for each source task. Furthermore, an object-dependent tolerance area around each target is defined, in which the reward is collected. For example, a throwing task is sequenced into two source tasks:
\paragraph{First source task:} The object is the hand; the achieved goal is the hand position; the desired goal is the ball position; the target is the ball, and the radius of the ball defines the threshold for the reward function. By manipulating the hand, the goal can be reached (Fig.\ref{fig:subtask1}).
\paragraph{Second source task:} Only after the agent has succeeded to get the hand on the ball, the ball can be manipulated and thrown towards the target (black-hole). In this case, the object is the ball; the achieved goal is the ball position; the desired goal is the position of the black-hole; the target is the black hole, and the threshold for the reward function is defined by the radius of the black-hole (Fig.\ref{fig:subtask2}).

\subsection{Knowledge transfer between source tasks} \label{ss:Knowledge transfer between sub-tasks}
To leverage the knowledge learned in the previous source task, we use the learned policy (the actor-network) as the initial policy for the next source task. This requires adjusting the input dimensions of the state spaces in the neural networks for each source task without affecting the output. We add the new dimensions by adding new input neurons for which all weights are set to $0$. With this initialization, the new dimensions do not affect the policy and keep the changes as smooth as possible when moving from one source task to another. When the agent starts learning a new source task, it tries to solve the previous task and incorrectly expects the episode to terminate. The agent then starts exploring around the object and will move it with high probability. By affecting the object, the agent begins collecting useful experiences to learn the new source task. Note that to manipulate the object in the $n$-th source task, the agent must first solve all prior $n-1$ source tasks (for example, to manipulate the ball, the agent must first reach it). In other words, every successful trajectory for the $n$-th source task encapsulates successful trajectories of all previous $n-1$ source tasks, and thus, preserves all previous policies. However, the exploration around the current object may also lead to episodes where the agent misses the object.  In this case, a set of positive virtual rewards are generated by HER, even though the agent did not complete all previous source tasks (see \cite{manela2019bias}). By encountering too many of these misleading samples, the agent may 'forget' what it has learned in prior source tasks. For this reason, Filtered-HER is applied \cite{manela2019bias}, which removes these bad samples. In section \ref{s:Experiments} we compare the performance of SHER with and without Filtered-HER.

\subsection{Implementation}\label{ss:implementation}
To train all the source tasks on the same simulation, we define the procedure \textit{state\_to\_obs}. This procedure gets the current state of the environment (the current observation concatenated with the goal) and returns the relevant observation dictionary, a reduced version of the target task. 
Before starting the learning process, we call the function \textit{get\_curriculum} (see algorithm \ref{alg:get_curriculum}) that returns a predefined list of \textit{state\_to\_obs} procedures for all the source tasks. While training, we call the current \textit{state\_to\_obs} procedure at each step, and the agent acts according to the returned observation. See algorithm \ref{alg:state_to_obs} for the pseudo-code.

\begin{algorithm}
	\caption{Build observation dictionary for source task}
	\label{alg:state_to_obs}
	\begin{algorithmic}[1]
		\Require
		\Statex $obs\_idx$ - indices for the observation
		\Statex $achieved\_idx$ - indices for the achieved goal
		\Statex $desired\_idx$ - indices for the desired goal
		\Input $S$ (current state) - The real observation concatenated with the real goal
		\Output $modified\_obs$ - The modified $obs$ dictionary
		\Procedure{state\_to\_obs}{$S$}
		\State $modified\_obs[observation] = S[obs\_idx]$
		\State $modified\_obs[achieved\_goal] = S[achieved\_idx]$
		\State $modified\_obs[desired\_goal] = S[desired\_idx]$
		\EndProcedure
	\end{algorithmic}
\end{algorithm}

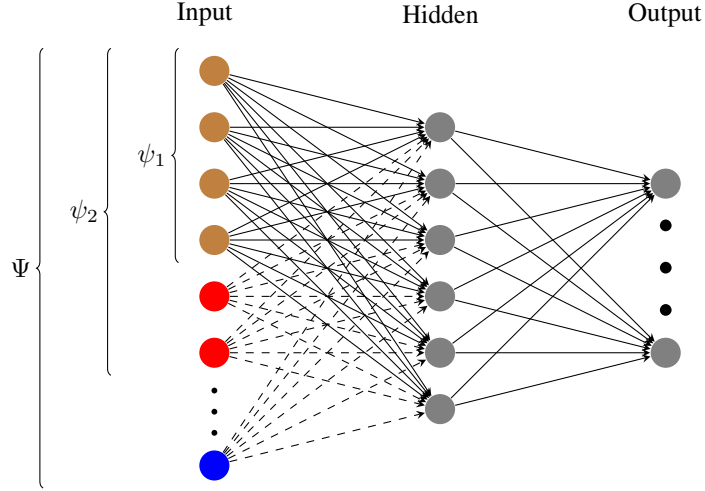
\begin{figure}
	\centering	
	\tikzset{%
		neuron missing1/.style={
			draw=none, 
			scale=2,
			text height=0.2cm,
			execute at begin node=\color{black}$\vdots$
		},
		neuron missing2/.style={
			draw=none, 
			scale=4,
			text height=0.2cm,
			execute at begin node=\color{black}$\vdots$
		}
	}
	\begin{tikzpicture}[x=1.5cm, y=1.5cm, >=stealth]
	\node[text width=1cm] at (0,2.5){Input};
	\foreach \m/\l [count=\y] in {1,...,4}
	{
		\node [circle,fill=brown,minimum size=0.4cm] (input-\m) at (0,2.5-0.5*\y) {};
	}
	
	\foreach \m/\l [count=\y] in {5,...,6}
	{
		\node [circle,fill=red,minimum size=0.4cm] (input-\m) at (0,2.5-0.5*\m) {};
	}
	
	\node [neuron missing1] at (0,-1.1) {};
	
	\foreach \m/\l [count=\y] in {8}
	{
		\node [circle,fill=blue,minimum size=0.4cm] (input-\m) at (0,2.5-0.5*\m) {};
	}
	
	\node[text width=1cm] at (2,2.5){Hidden};
	\foreach \m [count=\y] in {1,...,6}
	\node [circle,fill=black!50,minimum size=0.4cm ] (hidden-\m) at (2,2-0.5*\y) {};

	\node[text width=1cm] at (4,2.5){Output};
	\foreach \m [count=\y] in {1}
	\node [circle,fill=black!50,minimum size=0.4cm ] (output-\m) at (4,1.0) {};
	
	\foreach \m [count=\y] in {2}
	\node [circle,fill=black!50,minimum size=0.4cm ] (output-\m) at (4,-0.5) {};
	
	\node [neuron missing2] at (4,0.1) {};
	
	\foreach \l [count=\i] in {1,...,8}
	\foreach \l [count=\i] in {1,...,6}
	\node [above] at (hidden-\i.north){};
	\foreach \l [count=\i] in {1,n}
	\node [above] at (output-\i.east){};
	
	\foreach \i in {1,...,4}
	\foreach \j in {1,...,6}
	\draw [->] (input-\i) -- (hidden-\j);
	
	\foreach \i in {1,...,6}
	\foreach \j in {1,...,2}
	\draw [->] (hidden-\i) -- (output-\j);
	\foreach \i in {5,6,8}
	\foreach \j in {1,...,6}
	\draw [dashed,->] (input-\i) -- (hidden-\j);
	
	\draw[decoration={brace,mirror,raise=5pt},decorate]
	(-1.4,2.2) -- node[left=6pt] {$\Psi$} (-1.4,-1.7);
	\draw[decoration={brace,mirror,raise=5pt},decorate]
	(-0.2,2.2) -- node[left=6pt] {$\psi_1$} (-0.2,0.3);
	\draw[decoration={brace,mirror,raise=5pt},decorate]
	(-0.8,2.2) -- node[left=6pt] {$\psi_2$} (-0.8,-0.7);
	\end{tikzpicture}
	
	\caption[Neural network architecture for curriculum learning]{Neural network architecture for curriculum learning.
	Brown neurons are input neurons for the first source task ($\psi_1$). The red and blue neurons are the neurons added for the second and $n$-th source tasks, respectively. The dashed weights are initialized with zeros. The number of output neurons depends on the algorithm used (e.g., DQN, DDPG). Our tasks consist of two source tasks.}
	\label{fig:neural network for curriculum}
\end{figure}

The agent trains on each source task until it reaches a predefined level of expertise. We measure The agent's expertise with a moving average over the agent's success rate. When the agent reaches an average of $90\%$ success rate within a window size of $k$, the algorithm automatically switches to the next source task (see algorithm \ref{alg:learned_task}). We define task-overfitting, and policy's re-maneuverability as the agent's ability to adapt its policy to new tasks, and investigated it for different window sizes $k$ in section \ref{ss:experiments task overfitting}. When switching between source tasks, the weights of the new dimensions are initialized with zeros\footnote{For simplicity, all networks are implemented in their final full-size architecture and all weights, which are not used in the current source task, are initialized to zero (see Fig. \ref{fig:neural network for curriculum}). To match the sample's and networks' dimensions, we pad the samples with zeros. By doing so, the local gradients of the un-used weights are always zero, and the weights do not change.}. We compared different weight initialization methods for the critic network in section \ref{ss:experiments Networks Initialization Methods}. The experience buffer is emptied for each source task to only include samples relevant to the current source task. See algorithm \ref{alg:SHER} for the pseudo-code.

\newpage

\begin{algorithm}
	\caption{get curriculum}
	\label{alg:get_curriculum}
	\begin{algorithmic}[1]
		\Require
		\Statex $state\_to\_obs$ function for each source task
		\Statex padding vector for each source task
		\Statex reward function threshold for source task
		\Output $curriculum$ - A list containing all the information for each source task
		\Procedure{get\_curriculum}{}
		\State curriculum $\leftarrow$ []
		\For{each source task}
		\State task $\leftarrow$ dict()
		\State task[sto] $\leftarrow$ $state\_to\_obs$
		\State task[rft] $\leftarrow$ $reward\_function\_threshold$
		\State task[pad] $\leftarrow$ $padding$
		\State curriculum += task \Comment{+= denotes for append}
		\EndFor
		\EndProcedure
	\end{algorithmic}
\end{algorithm}

\begin{algorithm}
	\caption{learned task}
	\label{alg:learned_task}
	\begin{algorithmic}[1]
		\Require
		\Statex $c$ - success rate threshold \Comment{we used c = 90\%}
		\Statex $k$ - window size
		\Input $H$ - The success rate history
		\Output $learned$ - A Boolean. True if the performances reached the threshold
		\Procedure{learned\_task}{$H$}
		\State results $\leftarrow$ average($H$[$end-k:end$]) \Comment{last $k$ samples}
		\State $learned \leftarrow results \ge c$
		\EndProcedure
	\end{algorithmic}
\end{algorithm}

\begin{algorithm}
	\caption{Sequential-HER (SHER)}
	\label{alg:SHER}
	\begin{algorithmic}[1]
		\Require
		\Statex \textbullet~ an off-policy RL algorithm $\mathbb{A}$,\Comment e.g. DQN, DDPG 
		
		\State Initialize $\mathbb{A}$
		\State curriculum $\leftarrow$ get\_curriculum() \Comment{See algorithm \ref{alg:get_curriculum}}
		\State Initialize unused weights to zero
		
		\For{each $task$ in curriculum}
		\State $pad, sto, r\!ft \leftarrow task[pad],\, task[sto],\, task[r\!ft]$ \footnotemark[1]
		\State Initialize replay buffer $R$,\, success rate history $H$
		\State $learned$ $\leftarrow$ False
		\While{$learned$ is false}
		\For{$episode \leftarrow 1, M$}
		\State Sample a goal $g$ and an initial state $s_0$.
		\State $obs \leftarrow sto(s_0||g)$ \footnotemark[2] \Comment see algorithm \ref{alg:state_to_obs}
		\For{$t \leftarrow 0, T-1$}
		\State $\hat{s_t} \leftarrow obs[observation]$ ,\, $\hat{g_t} \leftarrow obs[desired\_goal]$
		\State Sample an action $a_t$ using the behavioral policy from $\mathbb{A}$: \par
		$\qquad \qquad a_t \leftarrow \pi(\hat{s_t}||\hat{g}||pad)$ \Comment hat stands for modified state / goal
		\State Execute the action $a_t$ and observe a new state $s_{t+1}$
		\State obs $\leftarrow sto(s_{t+1}||g)$
		\EndFor
		\EndFor
		\State Perform filtered-HER \Comment{see \cite{manela2019bias}}
		\State $learned$ $\leftarrow$ learned\_task($H$)  \Comment{see algorithm \ref{alg:learned_task}}
		\EndWhile
		\EndFor
	\end{algorithmic}
\end{algorithm}
\footnotetext[1]{$sto$ = state\_to\_obs, $r\!ft$ = reward\_function\_threshold}
\footnotetext[2]{|| denotes concatenation}
  
\subsection{Time Complexity of Sequential-HER}\label{ss:Time Complexity of Sequential-HER}
Next, we study the time complexity of SHER. Let $\Psi$ be a task consisting of a sequence of $n$ sub-tasks $\{\psi_1, ..., \psi_n\}$, where each sub-task is an MDP. To complete the task $\Psi$, the agent needs to solve sequentially each sub-task, that is, $\psi_1 \rightarrow \psi_2 \rightarrow \dots \rightarrow \psi_n$. An example of a task with two sub-tasks is our hand-throwing task. First, the agent needs to grab the ball with the hand and then throw it towards the target. The agent cannot move the ball before grabbing it with the hand.
Let $\mathcal{O}(\psi_i)$ be the state-space complexity (which we denote in the following as \emph{complexity} for simplicity) of sub-task $\psi_i$ and $\mathcal{O}_t(\psi_i)$ the time complexity (run-time) of sub-task $\psi_i$. 
A task's complexity is the sum of all its sub-tasks' complexities:
\begin{equation}
	\label{eq:task complexity}
	\mathcal{O}(\Psi) = \sum_i\mathcal{O}(\psi_i)\,.
	\end{equation}
	As shown in \cite{whitehead1991complexity,koenig1993complexity}, when following a random policy, the time complexity of a task is an exponential function of its complexity:
	\begin{equation}
	\label{eq:task time complexity}
	\mathcal{O}_t(\Psi) = \exp(\mathcal{O}(\Psi))\,.
	\end{equation}
	Using equations (\ref{eq:task complexity}) and (\ref{eq:task time complexity}), the time complexity can be expressed by
	\begin{equation}
	\mathcal{O}_t(\Psi) = \exp(\sum_i\mathcal{O}(\psi_i)) = \prod_i \exp(\mathcal{O}(\psi_i)) = \prod \mathcal{O}_t(\psi_i)\,.
\end{equation}
When using a sparse reward function without HER, the agent needs to solve the task by following a random policy. Hence, the time complexity of a task is the product of time-complexities for each sub-task.
When using a curriculum, the time complexity of a task reduces to about the sum of time complexities for each sub-task, since the agent solves only one sub-task at a time. By applying HER to each sub-task $\psi_i$, we reduced its time complexity (approximately) to a polynomial function of its complexity, i.e.,
\begin{equation}
	\mathcal{O}_t(\Psi) \approx \sum_i \mathcal{O}_t(\psi_i)= \sum_i \mbox{poly}[\mathcal{O}(\psi_i)]\,.
\end{equation}
We refer to appendix \ref{AppendixA} for sketch of a proof.\\
	

		

		
		

\section{Experiments} \label{s:Experiments}
Our algorithms were implemented and validated on three ball-throwing environments with different levels of complexity.
\subsection{Environments} \label{environments}
\begin{figure}
	\centering
	\subfloat[Hand Environment]
	{\tcbox{\includegraphics[height=0.2\textwidth]{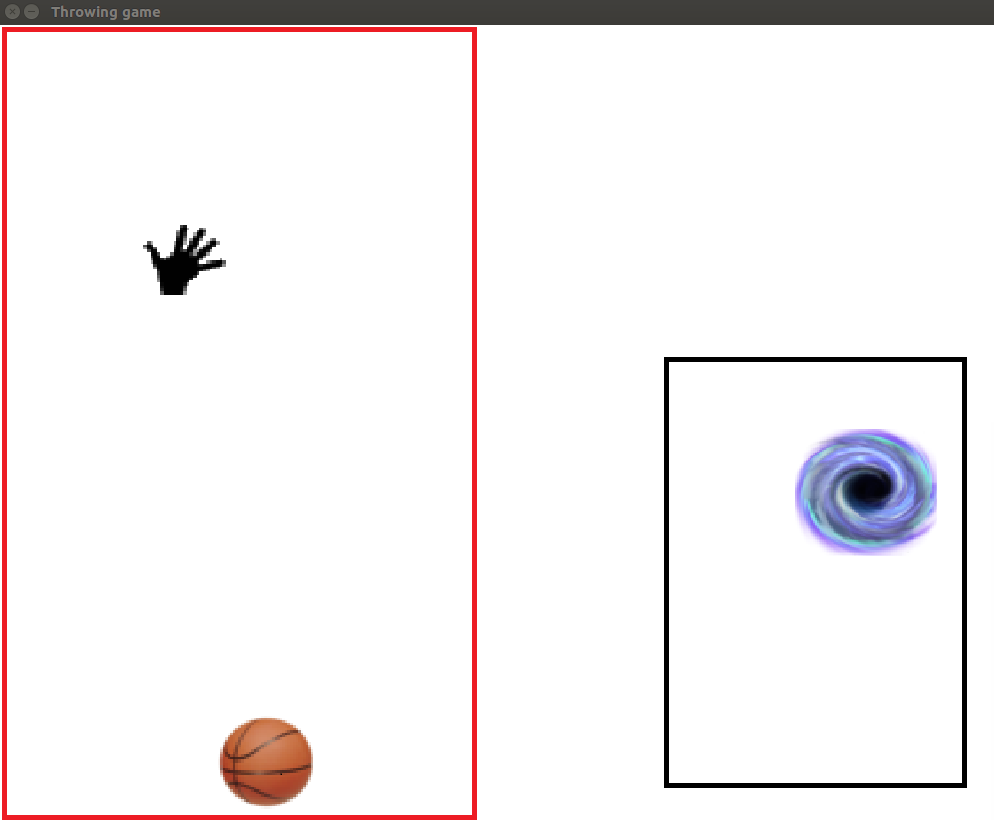}
	\label{fig:Hand-environment}}}
	\subfloat[Hand-Wall Environment]
	{\tcbox{\includegraphics[height=0.2\textwidth]{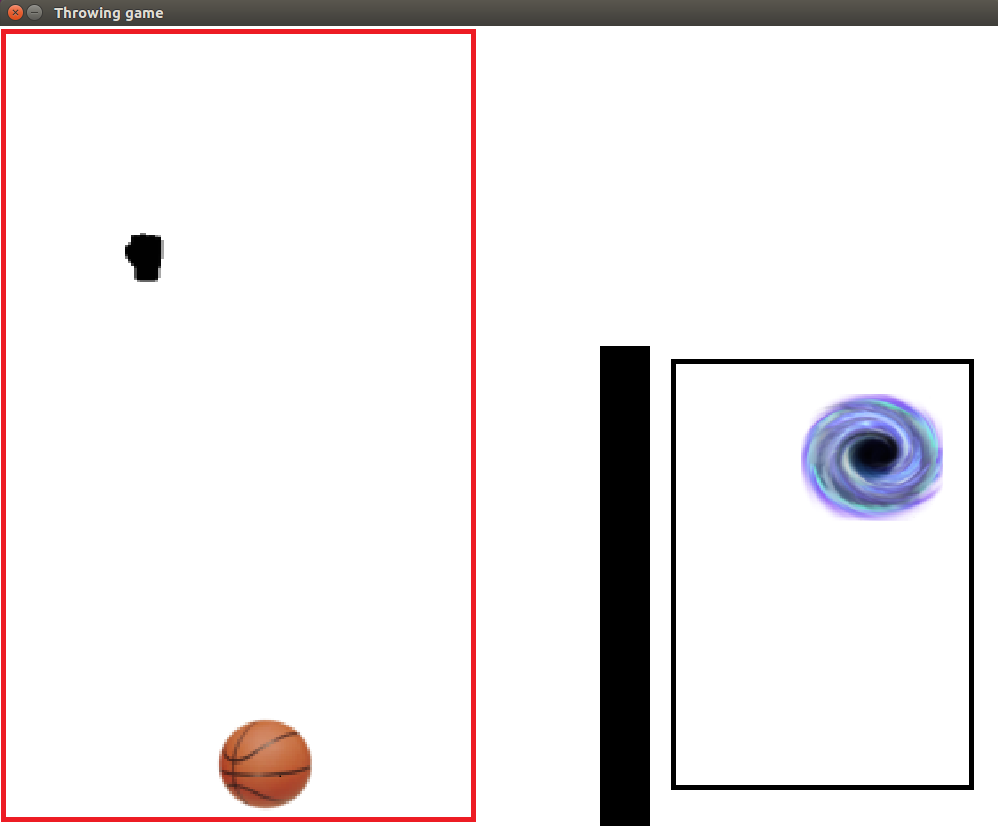} \label{fig:Hand-Wall-environment}}}%
	\subfloat[Robot Environment]
	{\tcbox{\includegraphics[height=0.2\textwidth]{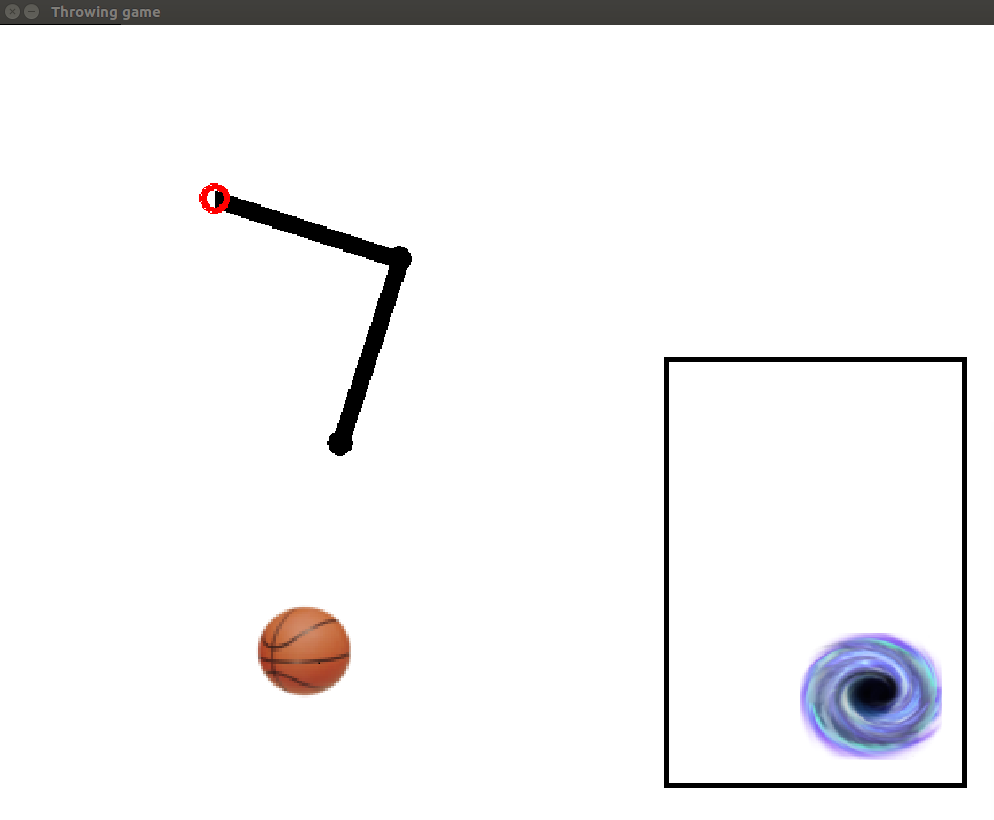} \label{fig:Robot-environment}}}%
		
	\caption{Environments. (a) \textit{Hand}: The agent needs to pick up the ball and throw it towards the black-hole. (b) \textit{Hand-Wall}: The agent needs to pick up the ball and throw it above the wall towards the black-hole. (c) \textit{Robot}: The agent needs to pick the ball with the end-effector by controlling joint velocities and throw it towards the black-hole. The red and black bounding boxes indicate the workspace of the hand and the region of possible target locations, respectively. The objects' dimensions were enlarged for visualization purposes.}
		\label{fig:environments}
\end{figure}

The environments were introduced in \cite{manela2019bias} and consist of the following tasks:
\begin{enumerate}
	\item \textbf{Hand:} In this task, the hand needs to pick up the ball and throw it at the target (Fig.\ref{fig:Hand-environment}).
	\item \textbf{Hand-Wall}: Same as the \textit{Hand} task, but in addition, a wall is placed in-between the agent's workspace and the target. The agent needs to throw the ball above the wall (Fig.\ref{fig:Hand-Wall-environment}).
	\item \textbf{Robot}: In this task, a manipulator needs to pick up the ball and throw it at the target. The agent controls the end-effector via the joint velocities, similar to real-world scenarios (Fig.\ref{fig:Robot-environment}).
	\end{enumerate}
In all environments, the agent receives a binary reward: 0 if the target is achieved and -1 otherwise. In the original environments, introduced in \cite{manela2019bias}, the ball is initialized within the hand with a probability of $p=0.5$ to simplify the task. To emphasize the sequentiality of the tasks, we set $p$ to $0$; thus, the agent has first to learn how to pick the ball before it can throw it.
    
\subsection{Algorithms Performances} \label{Performances}
To test the performance of the algorithms, we ran on each environment the following versions: \textbf{vanilla-HER}, \textbf{Filtered-HER}, \textbf{Unfiltered-SHER}\footnotemark, \textbf{SHER}. In all algorithms we used Instructional-Based Strategy (IBS) \cite{manela2019bias} as a virtual-goal selection strategy in HER and prioritized experience replay (PER) \cite{schaul2015prioritized} for experience replay.
\footnotetext{As explained is section \ref{ss:Knowledge transfer between sub-tasks}\,, SHER uses Filtered-HER by default.\label{refnote}}
The algorithms' results are evaluated using two criteria: (i) success rate and (ii) non-negative rewards.
The first criterion evaluates the performances of the agent. The second criterion shows the total number of useful samples collected by the agent so far. Useful samples denote samples for which the reward is $0$ (success) \textit{and} the object (e.g., the ball) moved. In other words, these are samples from which the agent can learn how to progress in the desired task (as shown in \cite{manela2019bias}, positive virtual samples, for which the object did not move, are useless for learning and induce bias).
  
\subsubsection{Success Rate}
\begin{figure}
	\centering
	\subfloat[]{{\includegraphics[width=15.5cm]{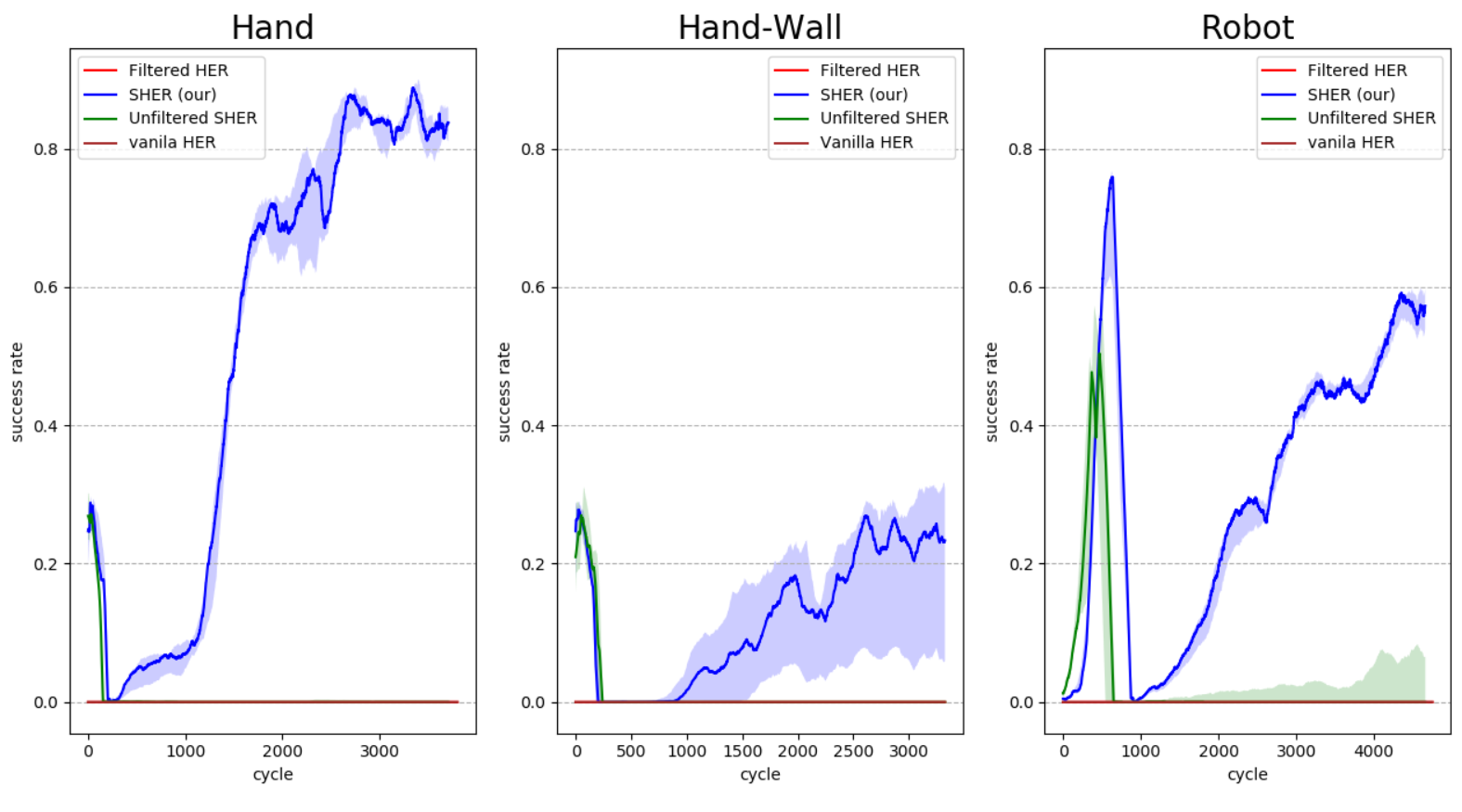} 
	\label{fig:Success rate}}} \\
	\caption{Learning curves for the multi-goal tasks. Results are shown over 15 independent runs. The bold line shows the median, and the light area indicates the range between the 33rd to 67th percentile.}
	\label{fig:performance}%
\end{figure}

As shown in Figures \ref{fig:Success rate}, both the vanilla-HER and Filtered-HER algorithms were not able to solve the tasks, resulting in almost zero success rate. The sequential nature of the task makes it too difficult for the agent to learn. In contrast, SHER learns the first source task and then successfully solves the full task. The agent trains on the first source task until it achieves a given threshold for a given number of cycles. The drops in performances indicate the time where the agent moved to the next source task. Note that in Fig.\ref{fig:Success rate}, it may appear as if the agent starts in the first sub-task with performances higher than zero and switches to the next sub-task before reaching a score of 0.9. However, these are artifacts due to the smoothing operation that we applied to the graphs. Smoothing may seem only to affect the first source task because changes in performances are slower in the second source task.\\
Next, we studied the effect of the filter on SHER. Unfiltered SHER achieved similar results to SHER on the first source-task, but then, unlike SHER, the agent quickly forgets all previous knowledge and fails in the second source task. Note that SHER learns the first source-task and quickly uptakes learning for the next source-task.
    
\subsubsection{Non-negative Rewards}
\begin{figure}
	\centering 
	\includegraphics[width=\textwidth]{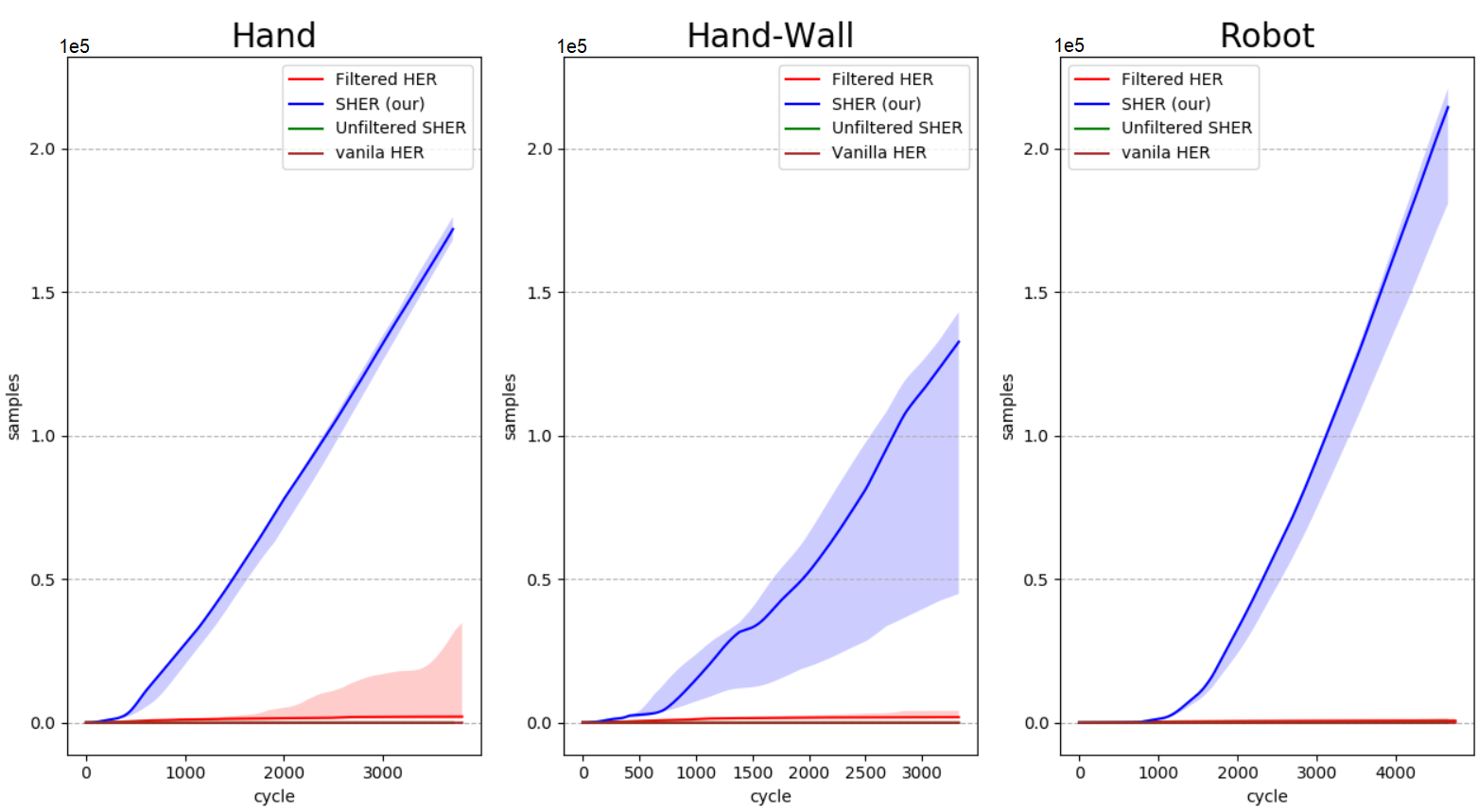}
	\caption{Positive rewards over time. The positive rewards are counted in tens of thousands. Results are shown over 15 independent runs. The bold line shows the median, and the light area indicates the range between the 33rd to 67th percentile. Samples are counted in tens of thousands.}
    \label{fig:Positive Rewards}
\end{figure}

A major problem of applying HER to sequential tasks results from the fact that the agent does not affect the achieved-goal, and therefore does not observe any useful experience. To learn, the agent must observe a sufficient amount of non-negative rewards (That is, in our case, rewards of 0). As shown in Fig. \ref{fig:Positive Rewards}, curriculum learning achieved in all tasks positive reward at a significantly faster rate than all other tested algorithms.
    
\subsection{Task Overfitting} \label{ss:experiments task overfitting}
\begin{figure}
	\centering
	\includegraphics[width=\textwidth]{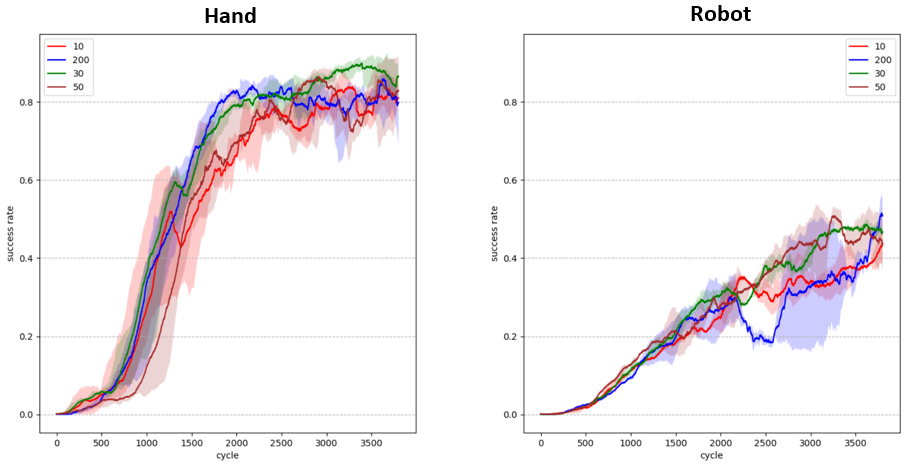}
	\caption{ Policy's re-maneuverability for different amounts of sub-task expertise. Results are shown over 10 independent runs. The bold line shows the median, and the light area indicates the range between the 33rd to 67th percentile.}
	\label{fig:Task Overfitting}
\end{figure}

SHER trains the agent on each source task to some level of expertise and then moves on to the next source task. The agent's knowledge is measured by a moving average on the success rate with a window size of $k$. We investigated next whether different window sizes affect the re-maneuverability of the policy due to task over- or under-fitting. To isolate the policy's adaptation time, we started counting cycles only after the agent solved the source-task and started learning the target task. As shown in Fig. \ref{fig:Task Overfitting}, the algorithm is relatively robust to different window sizes, and the difference in success rate is negligible. Nevertheless, we use a window size of 30, since it consistently showed a slight advantage, and is faster to achieve compared to larger window sizes.
    
\subsection{Networks Initialization Methods} \label{ss:experiments Networks Initialization Methods}
As described before, we initialize the new dimensions of the actor-network with all weights set to zeros. By initializing the new dimensions with zeros, we maintain the policy's updates as smooth as possible. In this section, we investigate the effect of different initialization methods for the critic network on the \textit{Hand} task. We first initialize the new weights regularly, i.e., the same initialization as the rest of the weights,  and then multiply the values by a factor $\alpha$. We compared three different settings:
(i) \textit{regular}: $\alpha = 1$, (ii) \textit{decreased weights}: $\alpha = 0.1$ (iii) \textit{reset weights}: $\alpha = 0$\,.
\begin{figure}
	\centering
	\includegraphics[width=0.7\textwidth]{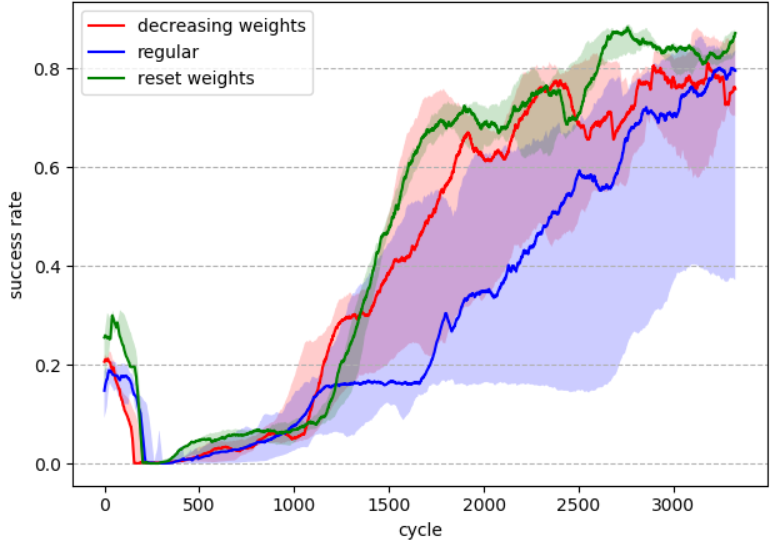}
	\caption{Success rate for different initialization methods in the \textit{Hand} task. Results are shown over 10 independent runs. The bold line shows the median, and the light area indicates the range between the 33rd to 67th percentile.}
	\label{fig:Networks Initialization Methods}
\end{figure}
As shown in Fig. \ref{fig:Networks Initialization Methods}, we obtained best results for the \textit{reset weights} setting. This choice implies that the critic function is not modified when adding more dimensions.

\section{Related Work}
Curriculum learning in reinforcement learning has been used in RL for several years but has recently received increased attention as RL agents are required to learn more complex tasks.
Like students in high-school, who study a curriculum of mathematics courses with an increasing level of complexity (e.g., arithmetics, algebra, analysis), an RL agent learns source tasks with increasing complexity levels to master a more difficult target task with better performance than learning it from scratch.
Different challenges can be addressed within curriculum learning. Selecting and sequencing suitable source tasks for the curriculum have been studied in \cite{narvekar2016source} using domain-dependent and domain-independent knowledge. The sequence of the curriculum is chosen by an expert.  Learning of curriculum policies, i.e., how to automatically choose the sequence of source tasks from a given set, has been presented in \cite{narvekar2017autonomous, narvekar2019learning}. As pointed out in \cite{narvekar2017curriculum}, further open questions of interest are the type of knowledge (e.g., value, policy, options, etc.) that can be transferred in the curriculum, as well as the transfer of learned curricula in one agent to different agents, similar to applying an existing curriculum to new students with different learning capabilities. The integration of a curriculum with deep reinforcement learning has been presented in \cite{ren2018self} for DQN by learning a curriculum for sampling the replay buffer. To put our work in context, we can identify two classes of algorithms of how to define the source tasks and select the curriculum.\\

\textit{Domain-independent curriculum:}
The first class of algorithms modifies the environment such that the source tasks are simpler versions of the target task while maintaining the main structure of the environment. These algorithms are domain-independent and do not require domain knowledge to design the source tasks. In \cite{saito2018curriculum} training starts with all states defined as goal-state and gradually focuses on the real goal-state. Another approach is to choose specific goals, which the agent can quickly achieve \cite{held2018automatic} or has already achieved \cite{andrychowicz2017hindsight}, and then use those to generalize to other goals. The application of a curriculum on the initial state is also possible.  As shown in \cite{florensa2017reverse}, training starts with an initial state closer to the target. However, this algorithm requires control over the manipulated object at every time step, and thus, it could not be applied to a throwing task. Another approach of manipulating the initial state was taken in \cite{salimans2018learning} using a single recorded trajectory played by an expert and set the initial state to states along the given trajectory. However, this method cannot be applied to multiple goals scenarios, as studied in this paper. Algorithms within this class mostly generate source tasks automatically with no need for expert knowledge. A related approach was also proposed by \cite{levy2017learning}, in which a target task (e.g., a reaching task over a large extent) is divided into a set of \textit{similar} sub-tasks (e.g., shorter reach tasks) and solving each sub-task in parallel using hierarchical reinforcement learning. Yet, this method solves a single long task rather than a sequential task consisting of multiple source tasks as considered here.\\\\
\textit{Domain-dependent curriculum:}
This class of algorithms trains the agent on fundamentally different source tasks by modifying the structure of the environment to provide the agent with an essential knowledge needed to learn the target task \cite{narvekar2019learning, narvekar2017autonomous}. These algorithms require domain knowledge to design the source tasks. For example, \cite{narvekar2019learning} learns how to play Ms. Pac-Man by first playing with no ghosts present, and then gradually introduces different types of ghosts, or \cite{narvekar2017curriculum} learns how to play chess with only a subset of chess pieces present. Most curriculum approaches from this class rely on the ability to provide the agent with specified source-tasks beforehand as well as source-task adjusted simulation environments to train the agent.
	
Our approach combines the two classes by splitting the full task into sub-tasks (source tasks) and simplifying each sub-task using HER. The sub-tasks and their order (curriculum) drop out naturally from the recurrent structure inherent in many object manipulation tasks. Like HER, our algorithm can be applied to multiple-goal scenarios. In contrast to existing algorithms from the second class, our method does not require an adjusted simulation for each sub-task. 
	
\section{Conclusion}
In this paper, we introduced a novel algorithm that enables using HER on sequential object manipulation tasks, called SHER. SHER applies HER on each source-task sequentially using a curriculum-learning approach. This approximately reduces the exponential time complexity of HER to polynomial time complexity in SHER. Furthermore, we presented a new technique that enables training the entire curriculum on the same simulation with no need for adjustments.
Using SHER, we have been able to train the agent on three different sequential object manipulation tasks with multiple goals and sparse reward functions, which has not been possible - to the best of our knowledge - with existing deep reinforcement learning algorithms.
	
\section*{Acknowledgement}
This research was supported in part by the Helmsley Charitable Trust through the Agricultural, Biological, and Cognitive Robotics Initiative and by the Marcus Endowment Fund both at Ben-Gurion University of the Negev. This research was supported by the Israel Science Foundation (grant no. 1627/17)
\newpage 
		
\bibliographystyle{unsrt}
\bibliography{references}
	
\newpage
\appendix
\section{Sequential HER - Time Complexity } \label{AppendixA}
\begin{figure}[H]
    \centering
    \tcbox{
    \begin{tikzpicture}[auto,node distance=8mm,>=latex,font=\small]
        \tikzstyle{round}=[thick,draw=black,circle]
        \node[round]  at (0, 0) (s0) {$s_0$};
        \node[round]  at (1.5, 0) (s1) {$s_1$};
        \node[round]  at (3, 0) (s2) {$s_2$};
        \node[round]  at (4.5, 0) (s3) {$s_3$};
        \node[round]  at (7, 0) (sn) {$s_n$};
        \node at ($(s3)!.5!(sn)$) {\ldots};
            
        \draw[->] (s0) to [out=315,in=225] (s1);
        \draw[->] (s1) to [out=315,in=225] (s2);
        \draw[->] (s2) to [out=315,in=225] (s3);
        \draw[->] (s1) to [out=135,in=45] (s0);
        \draw[->] (s2) to [out=135,in=45] (s1);
        \draw[->] (s3) to [out=135,in=45] (s2);
    
    \end{tikzpicture}
    }
    \caption[Toy problem]{Toy problem}
    \label{fig:toy problem}
\end{figure}
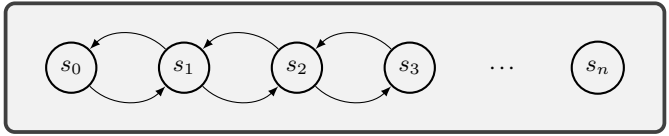

In this appendix, we sketch a proof for the time complexities of traditional RL algorithms, HER and Sequential-HER on tasks with a sparse reward function. For this purpose we consider the following toy problem: An agent needs to go from an initial state $s_0$ to a goal state $s_n$, which is $n$ steps away (Fig. \ref{fig:toy problem}).
\noindent For simplicity we make the following assumptions:
\begin{itemize}
    \item \textit{premise 1:} Under a random policy, the agent goes from state $s_i$ to $s_{i+1}$ with probability $p_i=p$.
    \item \textit{premise 2:} At each episode, the agent will first go to the best known state, and will continue with a random policy.
    \item \textit{premise 3:} The agent can learn how to reach a goal from a single successful example.
    \item \textit{premise 4:} There is a limited number of steps, namely $n$. Hence, the agent cannot waist moves.
\end{itemize}
    
\begin{figure}[H]
    \centering
    \tcbox{
    \begin{tikzpicture}[auto,node distance=8mm,>=latex,font=\small]
        \tikzstyle{round}=[thick,draw=black,circle]
        \node[round]  at (0, 0) (s0) {$s_0$};
        \node[round]  at (1.5, 0) (s1) {$s_1$};
        \node[round]  at (3, 0) (s2) {$s_2$};
        \node[round]  at (4.5, 0) (s3) {$s_3$};
        \node[round]  at (7, 0) (sn) {$s_n$};
        \node at ($(s3)!.5!(sn)$) {\ldots};
        
        \draw[->] (s0) to [out=315,in=225] (s1);
        \draw[->] (s1) to [out=315,in=225] (s2);
        \draw[->] (s2) to [out=315,in=225] (s3);
        \draw[->] (s1) to [out=135,in=45] (s0);
        \draw[->] (s2) to [out=135,in=45] (s0);
        \draw[->] (s3) to [out=135,in=45] (s0);
    \end{tikzpicture}
    }
    \caption[Toy problem - reset state]{Toy problem - reset state}
    \label{fig:toy problem - reset state}
\end{figure}
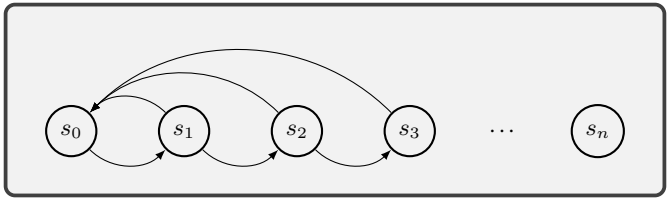

\subsection{Time complexity of traditional RL algorithms} \label{s:Time complexity of traditional RL algorithms}
Under \textit{premise 4}, the agent cannot make any mistakes. Thus, the game can be treated as a \textit{reset-state game} ( \cite{koenig1993complexity}). To solve the game, the agent must choose the right action at every state (Fig. \ref{fig:toy problem - reset state}). Without HER, the agent cannot leverage \textit{premise 2}, since all states seem equally bad. Under \textit{premise 3}, the agent needs to reach the goal once to learn the task. Under \textit{premise 1}, in each state, the agent goes forward with probability $p$. Hence, the probability of the agent to reach the goal is $p^n$. Thus, the time complexity of solving the task is $\frac{1}{p}^n$.
    
\subsection{Time complexity of HER} \label{s:Time complexity of HER}
When using HER, every visited state can be treated as a potential goal. Thus, under \textit{premise 3}, the agent memorizes the states it has seen and estimates how good this state is. Under \textit{premise 2}, the agent can then reach the best-known states (closest to the goal), and start exploring from there. Therefore, to reach the goal, the agent can learn state by state. Thus, the time complexity of solving the task using HER is the sum of the time complexities of all states: $\sum_{i=1}^n \frac{1}{p_i}$. That is, under \textit{premise 1}, $n\cdot\frac{1}{p}$.

\begin{figure}[th]
	\centering
	\tcbox{
		\tikzset{%
			three_dots/.style={
				draw=none, 
				scale=1,
				text height=0.2cm,
				execute at begin node=\color{black}$\vdots$
		}}
		
		\begin{tikzpicture}[auto,node distance=8mm,>=latex,
		brace/.style={decorate,decoration={brace,aspect=#1,amplitude=10pt}}]
		\tikzstyle{round1}=[thick,draw=black,circle, font=\large, minimum size=2pt]
		\tikzstyle{round2}=[thick,draw=black,circle, font=\small, minimum size=2pt]
		\tikzstyle{round3}=[thick,draw=black,circle, font=\scriptsize, minimum size=2pt]
		
		\node[round1]  at (0, 0) (s0) {$s_0$};
		\node[round1]  at (1.5, 0) (s1) {$s_1$};
		\node[round1]  at (3, 0) (s2) {$s_2$};
		\node[round1]  at (4.5, 0) (s3) {$s_3$};
		\node[round1]  at (6.5, 0) (sn) {$s_n$};
		\node at ($(s3)!.5!(sn)$) {\ldots};
		
		\draw[->] (s0) to [out=315,in=225] (s1);
		\draw[->] (s1) to [out=315,in=225] (s2);
		\draw[->] (s2) to [out=315,in=225] (s3);
		\draw[->] (s1) to [out=135,in=45] (s0);
		\draw[->] (s2) to [out=135,in=45] (s0);
		\draw[->] (s3) to [out=135,in=45] (s0);
		
		\draw [brace=0.1875] (0, -1) -- (4, -1);
		\node[round2]  at (0, -1.75) (s0_1) {$s_0$};
		\node[round2]  at (1.25, -1.75) (s1_1) {$s_1$};
		\node[round2]  at (2.5, -1.75) (s2_1) {$s_2$};
		\node[round2]  at (4, -1.75) (sn_1) {$s_n$};
		\node at ($(s2_1)!.5!(sn_1)$) {\ldots};
		
		\draw[->] (s0_1) to [out=315,in=225] (s1_1);
		\draw[->] (s1_1) to [out=315,in=225] (s2_1);
		\draw[->] (s1_1) to [out=135,in=45] (s0_1);
		\draw[->] (s2_1) to [out=135,in=45] (s0_1);
		
		\node[three_dots] at (0.625, -2.7) {};
		
		\draw [brace=0.1923] (0, -3.35) -- (3.25, -3.35);
		\node[round3]  at (0, -4) (s0_2) {$s_0$};
		\node[round3]  at (1, -4) (s1_2) {$s_1$};
		\node[round3]  at (2, -4) (s2_2) {$s_2$};
		\node[round3]  at (3.25, -4) (sn_2) {$s_n$};
		\node at ($(s2_2)!.5!(sn_2)$) {\ldots};
		
		\draw[->] (s0_2) to [out=315,in=225] (s1_2);
		\draw[->] (s1_2) to [out=315,in=225] (s2_2);
		\draw[->] (s1_2) to [out=135,in=45] (s0_2);
		\draw[->] (s2_2) to [out=135,in=45] (s0_2);
		
		\draw[decoration={brace,mirror,raise=5pt},decorate]
		(-2,0.5) -- node[left=6pt] {\large{$\Psi$}} (-2,-4.5);
		\draw[decoration={brace,mirror,raise=5pt},decorate]
		(-1,-1.25) -- node[left=6pt] {$\psi_{n-1}$} (-1,-4.5);
		\draw[decoration={brace,mirror,raise=5pt},decorate]
		(-0.3,-3.5) -- node[left=6pt] {$\psi_1$} (-0.3,-4.5);
		
		\end{tikzpicture}
	}
	\caption[Toy problem: sequential reset-state task]
	{Toy problem: sequential reset-state task.\\
		$\Psi$ stands for the full task. $\psi_1$ and $\psi_{n-1}$ stands for the first and one-before-last source tasks, respectively. To make any progress in each task (get from $s_0$ to $s1$), the agent must first solve all previous source tasks.}
	\label{fig:toy problem - sequential reset state}
\end{figure}
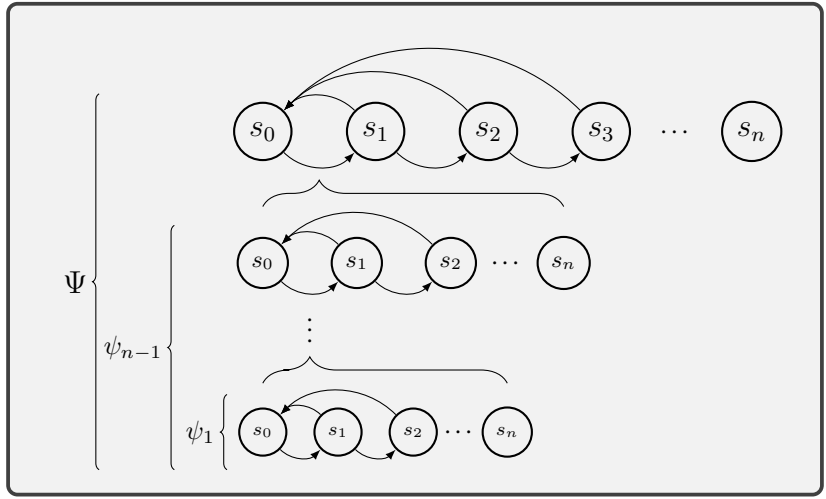

\subsection{Time complexity of Sequential-HER}\label{s:Time complexity of Sequential-HER}
In a sequential task, the agent must solve all source tasks before it can manipulate the object and make any progress (see Fig. \ref{fig:toy problem - sequential reset state}). For example, in our tasks, the agent must first reach the ball before it can move it. In other words, the time complexity of getting from $s_0$ to $s_1$
in the target task $\Psi$ is equal to the time complexity of all the source tasks $\psi_1,\dots \psi_{n-1}$. As HER in its original formulation is not applied to the source task level, but only on the target task, the time complexity of solving all source tasks is exponential in their total state-space complexity (see \ref{s:Time complexity of traditional RL algorithms}). By applying HER sequentially to each source-task, the tie complexity of getting from $s_0$ to $s_1$ of source task $\psi_i, i=1,\dots, n$ is instead polynomial in the total state-space complexity of former source-tasks $\psi_1,\dots, \psi_{i-1}$. (\ref{s:Time complexity of HER}).

\newpage
\section{Experiment Details} \label{appndx:B}
In this appendix, we provide a description of the experimental details, including networks' architectures and hyper-parameters.

\subsection{Training algorithm}
All the training was done using the DDPG algorithm with the following parameters:
\begin{center}
	\begin{tabular}{||c|c||} 
		\hline
		hyper-parameters & value \\ [0.5ex] 
		\hline\hline
		discount factor ($\gamma$) & 0.98\\ 
		\hline
		target-networks smoothing ($\tau$) & 7 \\
		\hline
		buffer size & 1e6 \\
		\hline
		$\epsilon$ initial value & 1 \\
		\hline
		$\epsilon$ decay rate & 0.95 \\
		\hline
		$\epsilon$ final value & 0.05 \\
		\hline
	\end{tabular}
\end{center}
For exploration we used a decaying epsilon-greedy policy:

\begin{equation*}
a =
\begin{cases*}
a^{*} & with probability $1-\epsilon$ \\
a^{*}+\mathcal{N}(0,\,I\cdot\sigma)   & with probability $0.8\cdot\epsilon$ \\
rand(a) & with probability $0.2\cdot\epsilon$
\end{cases*}
\end{equation*}
Where $\sigma=0.05\cdot action\_range$ and $\epsilon$ decays at the beginning of every epoch.\\
For experience replay we used \textit{prioritize experience replay} \cite{schaul2015prioritized}.
\subsection{Neural networks}
We used the same neural network layout for all the experiments:
\subsubsection{Actor:}
\begin{center}
	\begin{tabular}{||c|c|c|c|c|c||} 
		\hline
		layer & size & type & activation & BN  & additional info\\ [0.5ex] 
		\hline\hline
		input&input dim&Input&relu&No&No \\
		\hline
		hidden 1&64&FC&relu&No&No \\
		\hline
		hidden 2&64&FC&relu&No&No \\
		\hline
		hidden 3&64&FC&relu&No&No \\
		\hline
		output&action dim&FC&tanh&No&No \\
		\hline
	\end{tabular}
\end{center}
\begin{center}
	\begin{tabular}{||c|c||} 
		\hline
		hyper-parameter & value \\ [0.5ex] 
		\hline\hline
		learning rate & 0.001 \\
		\hline
		gradient clipping & 3 \\
		\hline
		batch size & 64 \\ 
		\hline
	\end{tabular}
\end{center}

\subsubsection{Critic:}
\begin{center}
	\begin{tabular}{||c|c|c|c|c|c||} 
		\hline
		layer & size & type & activation & BN  & additional info\\ [0.5ex] 
		\hline\hline
		input&input dim&Input&relu&Yes&No \\
		\hline
		hidden 1&64&FC&relu&Yes&concat the layer to the action \\
		\hline
		hidden 2&64&FC&relu&Yes&No \\
		\hline
		hidden 3&64&FC&relu&Yes&No \\
		\hline
		output&1&FC&linear&Yes&No \\
		\hline
	\end{tabular}
\end{center}
\begin{center}
	\begin{tabular}{||c|c||} 
		\hline
		hyper-parameter & value \\ [0.5ex] 
		\hline\hline
		learning rate & 0.001 \\
		\hline
		gradient clipping & 3 \\
		\hline
		batch size & 64 \\ 
		\hline
	\end{tabular}
\end{center}

\end{document}